\documentclass[conference]{IEEEtran}
\IEEEoverridecommandlockouts
\usepackage{cite}
\usepackage{amsmath,amssymb,amsfonts}
\usepackage{graphicx}
\usepackage{textcomp}
\usepackage{xcolor}
\usepackage{xspace}
\usepackage{xurl}
\def\BibTeX{{\rm B\kern-.05em{\sc i\kern-.025em b}\kern-.08em
    T\kern-.1667em\lower.7ex\hbox{E}\kern-.125emX}}
\usepackage{mathtools}
\usepackage{multirow}
\usepackage{adjustbox}
\usepackage{subcaption}
\usepackage{tikz}
\usetikzlibrary{positioning}
\usepackage{hyperref}
\usepackage{caption}
\usepackage[french,boxed,vlined,linesnumbered,inoutnumbered,rightnl,algo2e]{algorithm2e}
\usepackage{algorithm}
\usepackage{mwe}
\usepackage{stfloats}
\usepackage{pifont}

\begin{document}
% -----------------------------
\title{Automated Energy-Aware Time-Series Model Deployment on Embedded FPGAs for Resilient Combined Sewer Overflow Management

\thanks{The authors gratefully acknowledge the financial support provided by the Federal Ministry for Economic Affairs and Climate Action of Germany for the RIWWER project (01MD22007C, 01MD22007H).}}
\author{
    \IEEEauthorblockN{
    Tianheng Ling\textsuperscript{1}\IEEEauthorrefmark{1}, 
    Vipin Singh\textsuperscript{2}\IEEEauthorrefmark{1},
    Chao Qian\textsuperscript{1},  
    Felix Biessmann\textsuperscript{2,3} and 
    Gregor Schiele\textsuperscript{1}
    }
    \IEEEauthorblockA{
    \textsuperscript{1} University of Duisburg-Essen, Duisburg, Germany \\
    \textsuperscript{2} Berlin University of Applied Sciences and Technology, Berlin, Germany \\
    \textsuperscript{3} Einstein Center for Digital Future, Berlin, Germany \\
    \IEEEauthorrefmark{1} These authors contributed equally to this work.
    }
}
\maketitle
% -----------------------------
\begin{abstract}
Extreme weather events, intensified by climate change, increasingly challenge aging combined sewer systems, raising the risk of untreated wastewater overflow. Accurate forecasting of sewer overflow basin filling levels can provide actionable insights for early intervention, helping mitigating uncontrolled discharge. In recent years, AI-based forecasting methods have offered scalable alternatives to traditional physics-based models, but their reliance on cloud computing limits their reliability during communication outages.
To address this, we propose an end-to-end forecasting framework that enables energy-efficient inference directly on edge devices. Our solution integrates lightweight Transformer and Long Short-Term Memory (LSTM) models, compressed via integer-only quantization for efficient on-device execution. Moreover, an automated hardware-aware deployment pipeline is used to search for optimal model configurations by jointly minimizing prediction error and energy consumption on an AMD Spartan-7 XC7S15 FPGA.
Evaluated on real-world sewer data, the selected 8-bit Transformer model, trained on 24 hours of historical measurements, achieves high accuracy (MSE 0.0376) at an energy cost of 0.370~mJ per inference. In contrast, the optimal 8-bit LSTM model requires significantly less energy (0.009~mJ, over 40$\times$ lower) but yields 14.89\% worse accuracy (MSE 0.0432) and much longer training time. This trade-off highlights the need to align model selection with deployment priorities, favoring LSTM for ultra-low energy consumption or Transformer for higher predictive accuracy.
In general, our work enables local, energy-efficient forecasting, contributing to more resilient combined sewer systems.
All code can be found in the GitHub Repository\footnote{\url{https://github.com/tianheng-ling/EdgeOverflowForecast}}.

\end{abstract}
% -----------------------------
% -----------------------------

% No more than 6 pages, Single-Blind
% conferencne link: https://isc2-2025.org/
% All visualization is in https://drive.google.com/file/d/1BTmDNlndRNZ0LGXdlUgerfXDMVjh55mD/view?usp=sharing

%%%%%%%%%%%%%%%%%%%%%%%%%%%%%%%%%%%%%%%%%%%%%%%%%%%%%%%%%%%%%%%%%%%%%%%
\section{Introduction}
\label{sec:introduction} 

The growing frequency and severity of extreme weather events caused by climate change present serious challenges to aging urban infrastructure~\cite{allard2021climate}. Combined sewer systems, which are still common in many historical cities, are particularly vulnerable. These systems carry stormwater and wastewater through shared pipelines, making them prone to overflow during heavy rainfall~\cite{van2022towards}. When overwhelmed, they discharge untreated sewage into the environment, posing significant ecological and public health risks~\cite{sojobi2022impact, hammond2021detection}. Many cities have installed overflow basins as temporary buffers to mitigate peak loads. However, without accurate forecasts of their filling levels, it remains difficult to initiate timely preventive measures such as flow diversion~\cite{baneerjee2024overflow}.

Traditional physics-based forecasting models are often costly to maintain and difficult to adapt to evolving environmental conditions~\cite{saddiqi2023smart}. In recent years, data-driven approaches, especially \emph{Deep Learning} (DL) models, offer scalable and adaptive alternatives for forecasting~\cite{rosin2021committee}. Previous studies~\cite{singh2024data, singh2025robust} have shown that DL models can accurately predict overflow basin levels even under irregular rainfall and incomplete data. However, most existing methods depend on cloud computing, which may become unreliable or unavailable during extreme weather, precisely when predictions are most urgently needed.

To support resilient combined sewer management, we revisit the installation context of urban sewer overflow basins and identify three key system-level requirements for forecasting:  
(1) Local inference capability during cloud disruptions to ensure timely response and reliability.  
(2) Long-term autonomous operation on battery power with minimal maintenance during main power outages.  
(3) Forecasting accuracy sufficient to trigger reliable alerts and actionable interventions.

To meet these requirements, we propose deploying DL time-series models on edge devices placed directly alongside level sensors in overflow basins. These platforms process sensor data locally and forecast future filling levels in real time. When necessary, they can trigger actuators or issue early warnings. For long-term operation in power-constrained environments, we leverage low-power \emph{Field Programmable Gate Arrays} (FPGAs) to accelerate model inference. However, this deployment setting introduces three key challenges: (i) capturing both short-term fluctuations and long-term trends from limited input history, (ii) fitting model architectures within tight hardware budgets (logic, memory, and compute), and (iii) balancing forecasting accuracy and energy consumption. 
To tackle these constraints, this study makes the following contributions:
\begin{itemize}
    \item  We enhance a Transformer-tailored deployment toolchain targeting AMD Spartan-7 FPGAs~\cite{ling2025deployment} to support \emph{Long Short-Term Memory} (LSTM) models, enabling broader architecture selection under varying resource constraints.

    \item We conduct a systematic comparison between LSTM and Transformer models using a multi-objective model configuration search via Optuna that jointly minimizes validation loss and energy consumption per inference.

    \item Using real-world sewer data, we deploy models on a battery-powered \emph{ElasticNode V5} platform with an \emph{AMD Spartan-7 XC7S15} FPGA. The selected 8-bit Transformer, trained on 24 hours of historical measurements, achieves higher forecasting accuracy (MSE 0.0376) at an energy cost of 0.370~mJ per inference, while the chosen 8-bit LSTM consumes over 40$\times$ less energy (0.009~mJ) but with 14.89\% lower accuracy (MSE 0.0432) and longer training time. These trade-offs offer practical guidance for model selection based on deployment priorities.
\end{itemize}

The remainder of this paper is structured as follows: Section~\ref{sec:related_work} reviews related literature. Section~\ref{sec:methodology} details the proposed on-device forecasting framework. Section~\ref{sec:results_eval} describes the experimental setup and evaluates the results. Section~\ref{sec:conclusion_future_work} concludes the paper and outlines directions for future work.

%%%%%%%%%%%%%%%%%%%%%%%%%%%%%%%%%%%%%%%%%%%%%%%%%%%%%%%%%%%%%%%%%%%%%%%%%%%%%%%%
\section{Related Work}
\label{sec:related_work}

This section reviews DL applications in sewer system management, approaches for efficient deployment of DL models on edge devices, and recent advances in FPGA-based acceleration for time-series analysis.

%============================================================
\subsection{Deep Learning for Sewer System Management}

DL has shown increasing promise for intelligent sewer system management. Recent studies have explored DL-based solutions for tasks such as pollutant estimation, water quality monitoring~\cite{nagpal2024optimizing}, energy-efficient process control~\cite{monday2024review}, and anomaly detection in sensor data~\cite{seshan2024lstmencoder}. In particular, Singh et al.~\cite{singh2024data, singh2025robust} demonstrated that LSTMs, Transformers, and other time-series DL models can effectively forecast overflow basin levels, even under irregular rainfall and degraded sensor conditions. However, these models typically rely on cloud computing, which require stable connectivity.

Meanwhile, the proliferation of \emph{Internet of Things} (IoT) infrastructure has raised concerns about energy autonomy and system resilience under communication constraints~\cite{alshami2024iot}. Many existing approaches stream raw sensor data to the cloud, making them vulnerable during network outages. These limitations motivate the need for local forecasting on edge device \cite{ling2024towards}. While this shift has been conceptually recognized, rare works have explored how to optimize DL models under hardware-level constraints in such critical infrastructure contexts.

%==================================================
\subsection{Efficient Deep Learning Models on Edge Devices}

Deploying DL models on edge platforms introduces strict constraints on compute, memory, and energy. Lightweight architectures such as LSTMs and Transformers have been widely adopted to meet these constraints while preserving time-series modeling capacity~\cite{shuvo2022efficient}. LSTMs are robust to noisy or sparse sequences~\cite{seshan2024lstmencoder}, while Transformers model long-range dependencies efficiently~\cite{wen2022transformers}.

However, even compact architectures may struggle to meet the tightest hardware budgets on ultra-constrained devices. To further reduce deployment cost, model quantization has emerged as a standard technique. It reduces memory footprint and enables fixed-point or integer-only computation\cite{jacob2018quantization}, which is significantly more efficient on edge hardware. While Post-Training Quantization (PTQ) provides simplicity, it often fails to preserve accuracy in small models or fine-grained forecasting tasks~\cite{jiang2022ptq}. Quantization-Aware Training (QAT), in contrast, simulates low-precision arithmetic during training, helping models adapt to limited bitwidth representations and preserving accuracy across various deployment targets~\cite{nagel2021white}. In this work, we explore how lightweight architectures and QAT can be jointly optimized under real deployment constraints.

%==============================================================
\subsection{FPGA Acceleration for Time-Series Forecasting}

Recent research has investigated DL deployment across diverse edge hardware, including \emph{Microcontrollers Units} (MCUs), ASICs, and FPGAs~\cite{zhang2023review}. FPGAs are particularly attractive due to their reconfigurability, energy efficiency, and ability to exploit fine-grained parallelism for application-specific acceleration. Moreover, FPGAs allow flexible control over numerical precision, enabling customized support for low-bitwidth integer quantization. These properties make them well-suited for time-sensitive and energy-constrained scenarios in smart infrastructure systems~\cite{negi2021fpga}.

To harness these advantages in practice, recent efforts have developed toolchains that automate the deployment of quantized DL models onto FPGAs. Ling et al.~\cite{ling2025deployment} introduced a fully automated deployment framework for quantized Transformer models on AMD
Spartan-7 FPGAs. Their pipeline integrates integer-only quantization, RTL simulation, hardware synthesis, and a hardware-aware multi-objective model configuration search via Optuna. Here, we extend this toolchain with LSTM support and perform a systematic comparison between LSTM and Transformer architectures, offering practical deployment guidance under real-world constraints.

%%%%%%%%%%%%%%%%%%%%%%%%%%%%%%%%%%%%%%%%%%%%%%%%%%%%%%%%%%%%%%%%%%%%%%%%%%%%%%%%
\section{Methodology}
\label{sec:methodology}

This section presents our forecasting framework, including the edge hardware, the selected DL models, and the automated deployment pipeline for hardware-aware optimization.

%===============================
\subsection{Edge Device}
\label{subsec:hardware}

\begin{figure}[!htb]
\vspace{-5pt}
  \centering
  \includegraphics[width=.85\columnwidth]{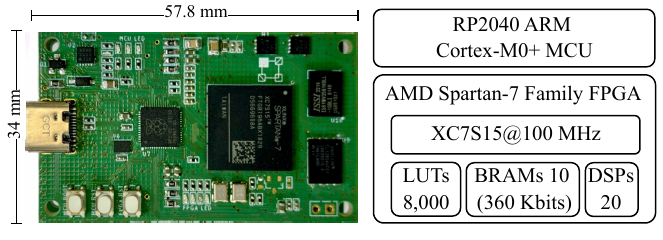}
  \caption{ElasticNode V5 Platform (adapted from \cite{ling2025deployment})}
  \label{fig:hardware_v5}
\vspace{-10pt}
\end{figure}

To satisfy the system-level requirements of local inference, long-term autonomy, and forecasting reliability outlined in Section~\ref{sec:introduction}, we adopt the \emph{ElasticNode V5} embedded platform introduced in~\cite{qian2023elasticai}. As shown in Figure~\ref{fig:hardware_v5}, this platform combines an \emph{RP2040 ARM Cortex-M0+} MCU with an \emph{AMD Spartan-7 XC7S15} (abbreviated as XC7S15) FPGA, offering 8,000 LUTs, 360 Kbits of BRAM, and 20 DSP slices running up to 741~MHz. To minimize energy usage, the MCU remains in low-power sleep mode by default (180~$\mu$A) and is periodically awakened by a timer to collect sensor data. The processed data is then transferred to the FPGA, which serves as a hardware accelerator and is powered only during model inference~\cite{qian2025configuration}. In this study, the system clock is fixed at 100~MHz.
%------------------------
%------------------------

%===============================
\subsection{Model Architectures}
\label{subsec:models}

\begin{figure}[!htb]
\vspace{-7pt}
    \centering
    \includegraphics[width=.75\columnwidth]{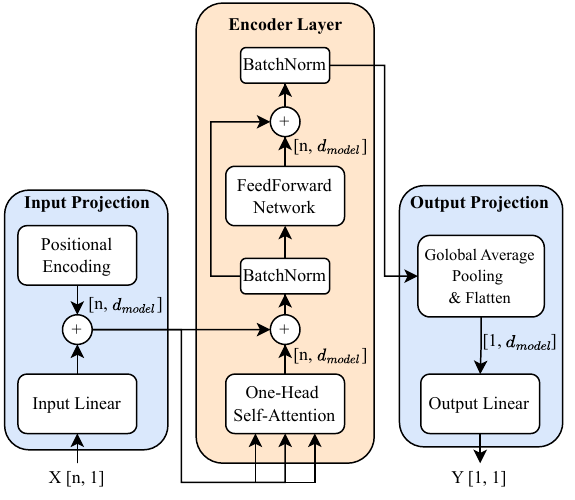}
    \caption{The Architecture of the Transformer Model\cite{ling2025deployment}}
    \label{fig:transformer_model}
\vspace{-5pt}
\end{figure}

To meet forecasting accuracy under hardware constraints, we adopt two compact DL architectures: an encoder-only Transformer and a single-layer LSTM. As shown in Figure~\ref{fig:transformer_model}, the Transformer model follows the lightweight, FPGA-friendly design proposed in~\cite{ling2024integer}. It comprises an input projection block, a simplified encoder layer, and an output projection block. To reduce resource usage and simplify training, all attention-related projections share a tunable embedding dimension $d_{\text{model}}$. The encoder layer includes a On-head Self-Attention module and a two-layer Feedforward Network with hidden dimension $4 \times d_{\text{model}}$. Additional implementation details are available in~\cite{ling2024integer}.

% ----------------------
% ----------------------

% ----------------------
\begin{figure}[!htbp]
\vspace{-5pt}
    \centering
    \includegraphics[width=0.9\columnwidth]{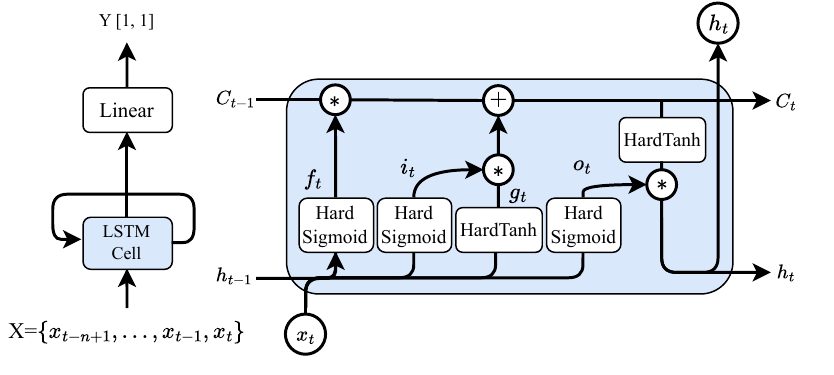}
    \caption{The Architecture of the LSTM Model}
    \label{fig:lstm_model}
\vspace{-5pt}
\end{figure}
% ----------------------

Figure~\ref{fig:lstm_model} shows the architecture of the LSTM model, consisting of a single vanilla LSTM layer followed by a linear layer. The model takes a univariate input sequence of fixed length ($n$) and outputs the predicted value for the next time step. Internally, the LSTM cell maintains temporal context using hidden and cell states, which are propagated across time steps. The hidden state at the final step is passed to the linear layer to generate the final prediction. The number of hidden units ($h_\text{size}$) is tunable to balance accuracy and deployability.
To ensure compatibility with FPGA resource constraints, we adopt the approach from~\cite{qian2024exploring}, replacing standard \emph{Sigmoid} and \emph{Tanh} activations with their hardware-friendly counterparts: \emph{HardSigmoid} and \emph{HardTanh}. These substitutions reduce arithmetic complexity and simplify non-linear function implementation. Further details on internal gating and activation approximations are available in~\cite{qian2024exploring}.

%=================================================
\subsection{Deployment Workflow with Hardware-Aware Optimization}
\label{subsec:deployment}

% ----------------------
\begin{figure*}[!htb]
    \centering
    \includegraphics[width=0.9\textwidth]{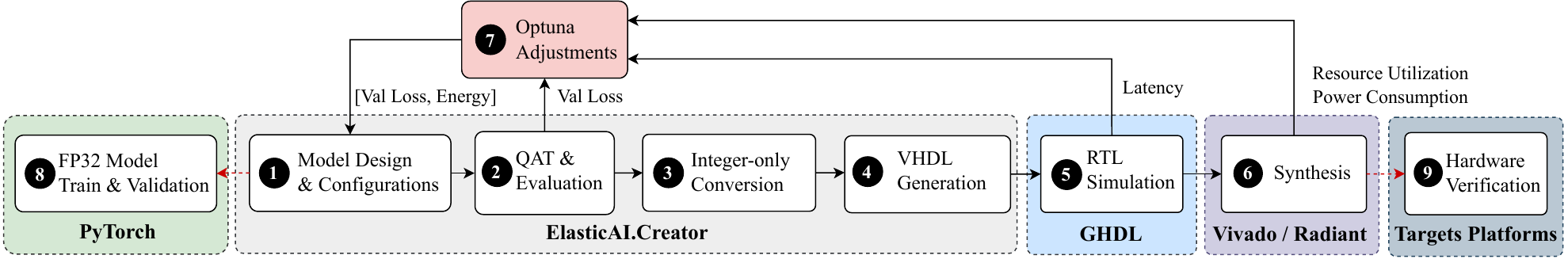}
    \caption{Overview of the Deployment Workflow, modified from ~\cite{ling2025deployment}}
    \label{fig:deployment_pipeline} 
\vspace{-15pt}
\end{figure*}
% ----------------------
To enable deployment of forecasting models on embedded FPGAs, we adopt the end-to-end deployment framework tailored for Transformer models proposed in~\cite{ling2025deployment}, which integrates QAT-based integer-only quantization (down to 4-bit), modular VHDL code generation, RTL simulation and FPGA synthesis. As displayed in Figure \ref{fig:deployment_pipeline}, this framework supports automated model translation from quantized models to hardware implementation and embeds real-time resource and power profiling into an Optuna-based hardware-aware model configuration search loop.

We extend this framework to support LSTM-based architectures. Specifically, we implement additional quantization modules and VHDL templates to support integer-only computation for LSTM-specific operations, including element-wise concatenation, \emph{Hadamard} product, \emph{HardTanh}, \emph{HardSigmoid}, LSTM cell, LSTM layer and LSTM block (i.e., stacked LSTM layers). Implementation details of the LSTM-specific quantization modules and VHDL templates are available in our open-source \emph{ElasticAI.Creator}\footnote{\url{https://github.com/es-ude/elastic-ai.creator/tree/add-linear-quantization}} library. These modules and templates follow the same quantization scheme as prior Transformer components. With this extension, both LSTM and Transformer models can be automatically quantized, synthesized, and evaluated on embedded FPGAs with energy, latency, and resource constraints enforced during search.

%%%%%%%%%%%%%%%%%%%%%%%%%%%%%%%%%%%%%%%%%%%%%%%%%%%%%%%%%%%%%%%%%%%%%%%%%%%%%%%%
\section{Experimental Results and Evaluation}
\label{sec:results_eval}

This section evaluates whether quantized LSTM and Transformer models meet the deployment requirements for the local, energy-efficient forecasting of overflow basin filling levels.We first describe the dataset and preprocessing method, then detail the hardware-aware search setting, and finally compare both models on the target FPGA.

%====================================
\subsection{Dataset and Preprocessing}

To enable a comparison with~\cite{singh2024data}, we use the same real-world sewer dataset collected by \emph{Wirtschaftsbetriebe Duisburg}\footnote{\url{https://www.wb-duisburg.de}}, the municipal utility of Duisburg, Germany. The dataset spans 2021–2023 and contains sensor measurements from the Vierlinden district, sampled at intervals ranging from 1 second to 1 hour. For local forecasting, we construct a univariate time series $\{x_t\}_{t=1}^{T}$ of hourly basin filling levels, yielding approximately 26,280 entries within a 0–6 meter range. The first two years of data were used for training and validation (80\%/20\% split), and the final year was reserved for testing.
Each model input consists of the past $n$ values $\{x_{t-n}, \dots, x_{t-1}\}$ used to predict the next value $x_t$. In contrast to~\cite{singh2024data}, which performs 12-step ahead forecasting by using 72 past observations, we adopt a simpler one-step ahead approach with $n \in \{6, 12, 24\}$ to capture short-, medium-, and long-term dependencies. This formulation better suits local scenarios requiring low-latency inference and continuous updates on resource-constrained devices. %This formulation better suits local forecasting scenarios, where low-latency inference and continuous real-time updates are prioritized over long-horizon prediction, especially under resource-constrained, on-device deployment settings.

%===============================================
\subsection{Hardware-Aware Hyperparameter Search}

We compare the deployment performance of quantized Transformer and LSTM models on the \emph{XC7S15} FPGA across different input lengths. For each setting, we conduct 100 Optuna trials using the \texttt{NSGAII} sampler to jointly minimize validation loss and energy consumption ($E \!=\! P \times T$) in millijoule (mJ). The validation loss is obtained during QAT and evaluation (\ding{182} in Figure \ref{fig:deployment_pipeline}), while the inference latency $T$ in milliseconds (ms) is measured from RTL simulation (\ding{186} in Figure \ref{fig:deployment_pipeline}). The post-synthesis power $P$ in milliwatts (mW) is extracted from Vivado 2019.2 reports after hardware synthesis (\ding{187} in Figure \ref{fig:deployment_pipeline}). 
Only configurations that satisfy all FPGA resource constraints (described in Section \ref{subsec:hardware}) are retained for the Pareto front, which consists of non-dominated configurations that represent optimal trade-offs between validation loss and energy consumption. The search space includes:

\begin{itemize}
    \item Quantization bitwidth: $b \in \{4, 6, 8\}$
    \item Batch size: $bs \in \{16, 32, \dots, 256\}$
    \item Learning rate: $lr \in [10^{-5}, 10^{-3}]$ (log-uniform)
    \item LSTM hidden size: $h_\text{size} \in \{8, 16, \dots, 64\}$
    \item Transformer model dimension: $d_\text{model} \in \{8, 16, \dots, 64\}$
\end{itemize}

Each trial runs for up to 100 training epochs with early stopping (patience \!=\! 10). Models are trained using the Mean Squared Error (MSE). All experiments use Python 3.11 with the Adam optimizer on an NVIDIA RTX 2080 SUPER GPU.% (CUDA 11.0). 

%====================================
\begin{figure*}[ht]
    \centering
    \begin{subfigure}[b]{0.3\textwidth}
        \centering
        \includegraphics[width=\linewidth]{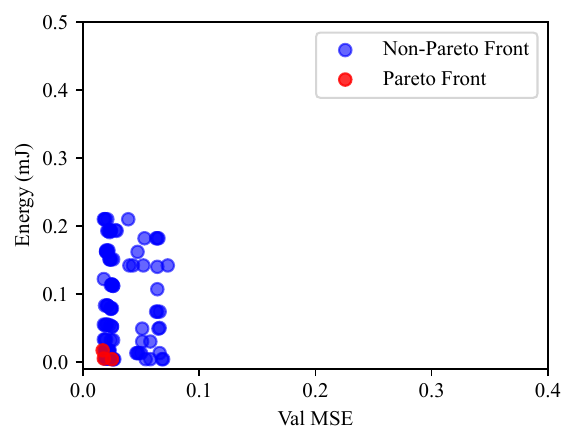}
        \caption{Transformer, $n\!=\!6$}
    \end{subfigure}
    \hfill
    \begin{subfigure}[b]{0.3\textwidth}
        \centering
        \includegraphics[width=\linewidth]{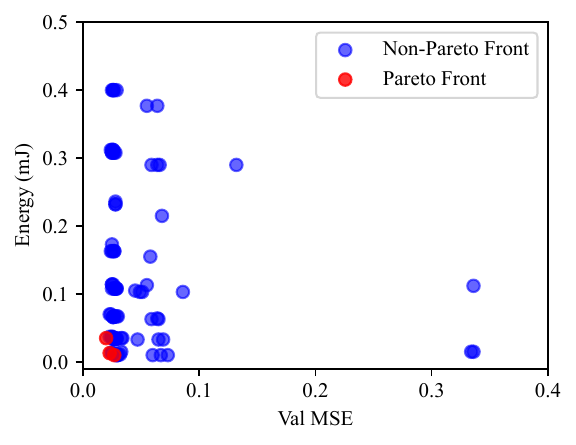}
        \caption{Transformer, $n\!=\!12$}
    \end{subfigure}
    \hfill
    \begin{subfigure}[b]{0.3\textwidth}
        \centering
        \includegraphics[width=\linewidth]{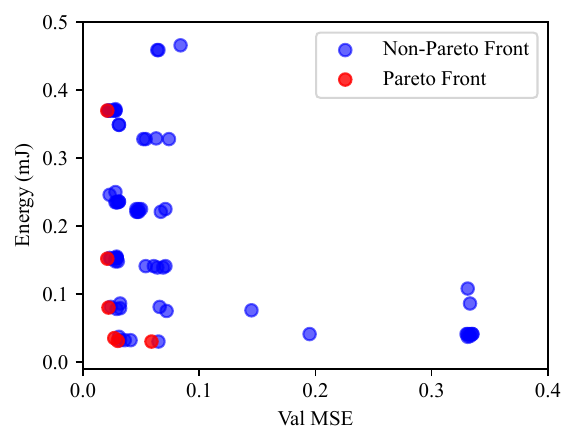}
        \caption{Transformer, $n\!=\!24$}
    \end{subfigure}
    \hfill
    \begin{subfigure}[b]{0.3\textwidth}
        \centering
        \includegraphics[width=\linewidth]{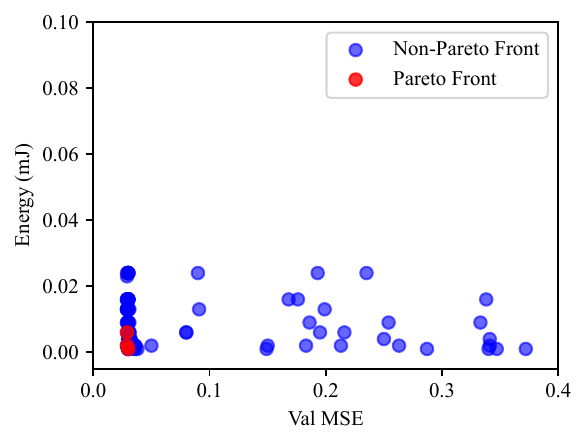}
        \caption{LSTM, $n\!=\!6$}
    \end{subfigure}
    \hfill
    \begin{subfigure}[b]{0.3\textwidth}
        \centering
        \includegraphics[width=\linewidth]{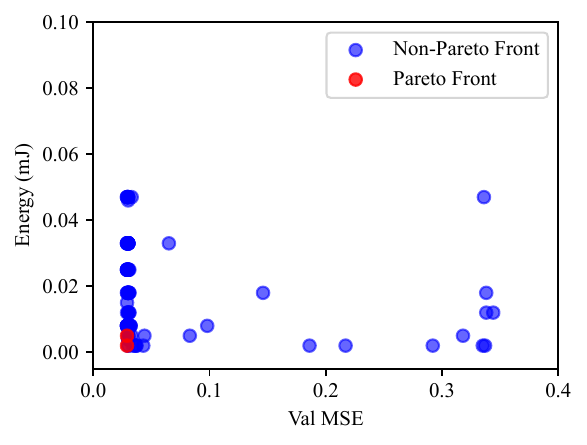}
        \caption{LSTM, $n\!=\!12$}
    \end{subfigure}
    \hfill
    \begin{subfigure}[b]{0.3\textwidth}
        \centering
        \includegraphics[width=\linewidth]{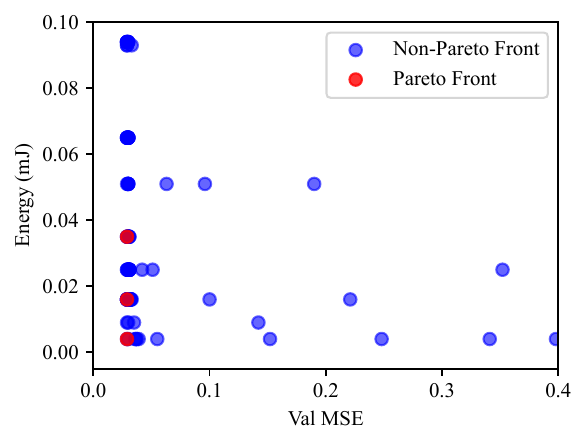}
        \caption{LSTM, $n\!=\!24$}
    \end{subfigure}
    \caption{Trade-offs between validation MSE and energy consumption (mJ) for Transformer and LSTM models on the XC7S15 FPGA. Each dot represents a deployable configuration, while optimal configurations at the Pareto Front are indicated in red.}
    \label{fig:pareto_all}
% \vspace{-10pt}
\end{figure*}

%==================================
\subsection{Deployment Results: Transformer}

Figure~\ref{fig:pareto_all}(a–c) show the Pareto fronts of Transformers obtained from hardware-aware search via Optuna. Each dot represents a deployable configuration, with red markers highlighting Pareto-optimal trade-offs between MSE and energy per inference. As input length increases, fewer configurations remain deployable due to rising model complexity and resource demand (100, 96, and 75 valid trials for $n \!=\! 6, 12, 24$, respectively). Moreover, the validation MSE distribution becomes more dispersed as the input length increases. Based on further analysis, this is partly because Optuna tends to favor smaller $d_\text{model}$ values to satisfy deployment constraints, which limits model complexity. Combined with the increased difficulty of learning long-term temporal patterns, this leads to greater variability in forecasting performance across trials.

%---------------------------
\begin{table*}[!htpb]
\centering
\caption{Selected Pareto-Optimal Model Configurations with Highest Integer-Only Test MSE on the \emph{AMD XC7S15} FPGA.}
\label{tab:optuna_results}
\resizebox{\textwidth}{!}{ 
\renewcommand{\arraystretch}{1.15}
\begin{tabular}{|c|c|c|c|c|c|c|c|c|c|c|c|c|c|c|}
\hline

\multirow{2}{*}{Model} & \multirow{2}{*}{$n$} & \multicolumn{4}{c|}{Configuration} & \multicolumn{3}{c|}{Test MSE} & \multirow{2}{*}{\begin{tabular}[c]{@{}c@{}}LUTs\\ (\%)\end{tabular}} & \multirow{2}{*}{\begin{tabular}[c]{@{}c@{}}BRAMs\\ (\%)\end{tabular}} & \multirow{2}{*}{\begin{tabular}[c]{@{}c@{}}DSPs\\ (\%)\end{tabular}} & \multirow{2}{*}{\begin{tabular}[c]{@{}c@{}}Energy\\ (mJ)\end{tabular}} & \multirow{2}{*}{\begin{tabular}[c]{@{}c@{}}Power$^{*}$\\ (mW)\end{tabular}} & \multirow{2}{*}{\begin{tabular}[c]{@{}c@{}}Latency$^{**}$\\ (ms)\end{tabular}} \\ \cline{3-9}
 & & b & bs & lr ($\times10^{-4})$ & $d_\text{model}$/$h_\text{size}$ & \multicolumn{1}{c|}{FP32} & Quantized & Variance (\%) &  & & & & & \\ \hline

\multirow{3}{*}{Transformer} &
6 & 6 & 80 & 3.653 & 8 & 0.0415 & 0.0427 & $\uparrow$2.89 & 43.56 & 15 & 95 & 0.004 & 49.0 & 0.091 \\ \cline{2-15} %trial6

&12 & 8 & 144 & 1.410 & 16 & 0.0394 & 0.0394 & $-$0.00& 55.88 & 100 & 100 & 0.036 & 67.0 & 0.532 \\ \cline{2-15} %trial56

&24 & 8 & 80 & 7.551 & 40 & 0.0339 & 0.0376 & $\uparrow$10.91 & 84.91 & 100 & 100 & 0.370 & 72.0 & 5.134 \\ \hline %trial79

\multirow{3}{*}{LSTM} &
6 & 8 & 112 & 1.210 & 16 & 0.0432 & 0.0433 & $\uparrow$0.23 & 32.84 & 0 & 55 & 0.002 & 48.0 & 0.046\\ \cline{2-15}

& 12 & 8 & 144 & 1.325 & 8 & 0.0433 & 0.0436 & $\uparrow$0.69 & 30.14 & 15 & 55  & 0.002 & 50.0 & 0.039 \\ \cline{2-15}

& 24 & 8 & 256 & 9.803 & 16  &  0.0434  & 0.0432 & $\uparrow$0.46 & 32.53 & 5 & 55   & 0.009 & 49.0 & 0.182\\ \hline %trial": 44,

\multicolumn{15}{l}{$n$ = input sequence length, $b$ = quantization bitwidth, $bs$ = batch size, $lr$ = learning rate, $d_\text{model}$ = Transformer's embedding dimension, $h_\text{size}$ = LSTM's hidden size.} \\
\multicolumn{15}{l}{$^{*}$Power was estimated at a temperature of 28.0$^{\circ}$C with a power deviation of 5.8\% compared to actual hardware measurements.} \\
\multicolumn{15}{l}{$^{**}$The latency obtained from GHDL compared to actual hardware measurements varies by about 2\%.}\\

\end{tabular}
}
\vspace{-15pt}
\end{table*}
%------------------
\begin{figure}[ht]
    \centering
    \begin{subfigure}[b]{0.493\columnwidth}
        \centering
        \includegraphics[width=\linewidth]{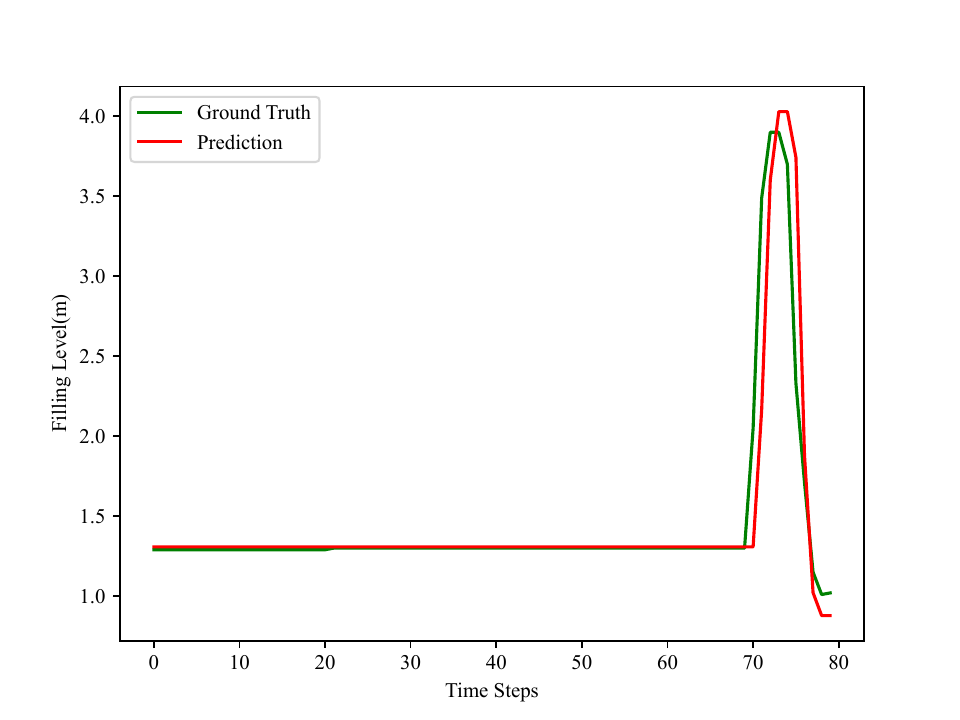}
        \caption{FP32 (MSE 0.0415)}
    \end{subfigure}
    \hfill
    \begin{subfigure}[b]{0.493\columnwidth}
        \centering
        \includegraphics[width=\linewidth]{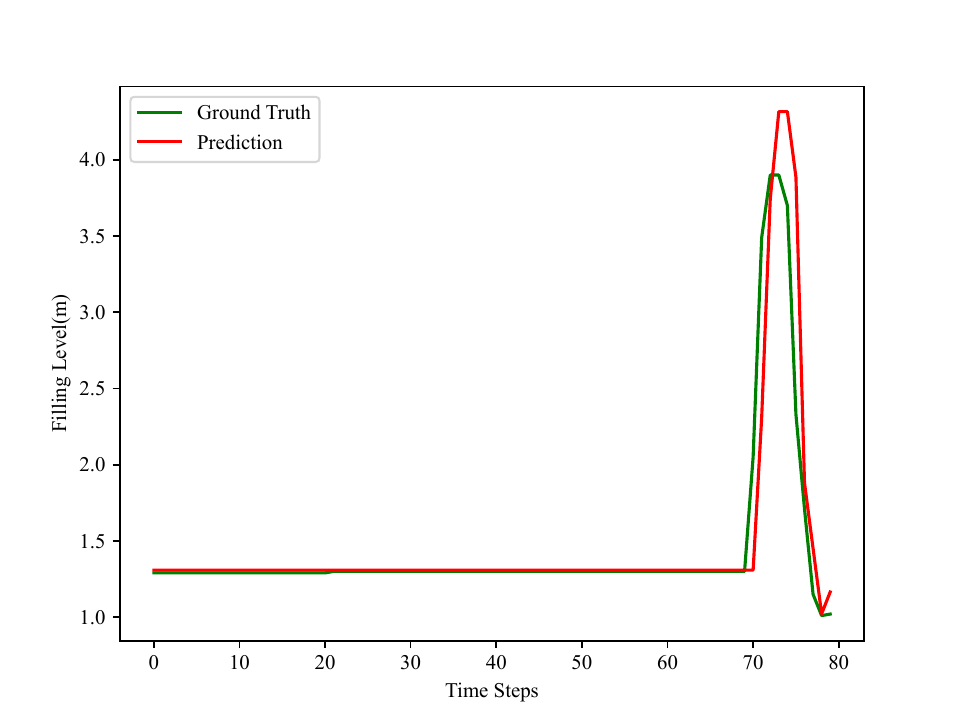}
        \caption{8-bit (MSE 0.0427)}
    \end{subfigure}
    \caption{Comparison between predicted (red) and actual (green) basin filling levels for the selected Transformer configuration ($n \!=\! 6$), showing accurate temporal prediction performance.}
    \label{fig:curves_transformer_6ws}
\vspace{-15pt}
\end{figure}
%------------------

The first three rows of Table~\ref{tab:optuna_results} present three Pareto-optimal Transformer configurations, 
selected based on the lowest test MSE achieved during integer-only inference. At $n \!=\! 6$, the selected model ($d_{\text{model}}\!=\!8$, 6-bit quantization) achieves a test MSE of 0.0427, just 2.89\% above FP32 baseline (0.0415), with an energy cost of 0.004 mJ and latency of 0.091 ms. Denormalized to the original 0–6 meter scale, this corresponds to a minor increase from 0.0627 to 0.0646 MSE, suggesting that the prediction accuracy remains practically sufficient. As visualized in Figure~\ref{fig:curves_transformer_6ws}(b), the quantized model closely tracks real basin filling trends, albeit with slightly reduced precision in capturing fine-grained variations compared to the FP32 model in Figure~\ref{fig:curves_transformer_6ws}(a).
As input length increases to 12 and 24, accuracy improves (down to 0.0376 MSE), but energy consumption grows steeply to 0.370 mJ (92.5$\times$ higher) and latency to 5.134 ms. Notably, when $n \!=\! 24$, the test MSE of the quantized model increases by 10.91\% relative to its FP32 counterpart, larger than the gap observed at shorter input lengths. However, despite this increase, it still achieves lower MSE than the quantized models with shorter input lengths. These results confirm the expected trade-off: longer sequences improve temporal modeling but incur significant energy and latency costs, making short input lengths a more viable option for energy-constrained deployments.
While our framework supports down to 4-bit precision, we observed that such configurations often result in unacceptable accuracy degradation for regression-based forecasting. Consequently, the Optuna-based search consistently favors 6- or 8-bit quantization, balancing model compactness with sufficient predictive performance.

%===========================
\subsection{Deployment Results: LSTM}

Figure~\ref{fig:pareto_all}(d–f) displays the Pareto fronts for LSTM models. All 100 trials are deployable across all input lengths ($n \in \{6, 12, 24\}$), and MSE distributions grow more dispersed as $n$ increases, similar to the Transformer trend. At each input length, LSTMs exhibit greater variance in MSE than Transformers. However, LSTM configurations consistently achieve low energy consumption, with nearly all trials under 0.1 mJ, well below the Transformer’s upper bound of 0.5 mJ.

As displayed in Table \ref{tab:optuna_results}, at $n \!=\! 6$, the selected 8-bit quantized LSTM achieves a test MSE of 0.0433, only a 0.23\% increase from its FP32 baseline, while consuming just 0.002 mJ per inference and running at 0.046 ms latency. Compared to the Transformer at the same input length (0.0427 MSE, 0.004~mJ), the LSTM yields 1.40\% higher MSE but consumes half the energy. In practical terms, this difference in test MSE may be negligible for overflow forecasting, especially when energy budget is the dominant concern. 
At $n\!=\!12$, the MSE increases marginally to 0.0436, yet energy and latency remain unchanged. Interestingly, this suggests that increasing input length does not significantly benefit LSTM’s performance, likely due to its limited capacity to extract longer-term temporal patterns compared to Transformer models. Specifically, after increasing the input sequence length to 24, the model improves in accuracy (0.0432 MSE) with energy rising to 0.009 mJ, still 41.1$\times$ lower than the Transformer’s 0.370 mJ. However, LSTM shows small accuracy gains when the input sequence length increases from 6 to 24 (0.0433 $\rightarrow$ 0.0432).

%=========================
% \vspace{-6pt}
\subsection{Discussion}

Despite these advantages, LSTM models incur a higher end-to-end deployment cost in terms of training and RTL simulation runtime. At $n\!=\!6$, the full deployment pipeline (as shown in Figure \ref{fig:deployment_pipeline}) for a Transformer model completes up to 1 hour. In contrast, the same pipeline for LSTM takes up to 2 hours due to the sequential nature of recurrent computations and the additional complexity of LSTM-specific hardware modules. This runtime disparity increases further with increasing input length, limiting the scalability of large-scale automated exploration for LSTM-based models. From an edge-computing perspective, where energy efficiency is often prioritized over marginal accuracy gains, LSTM offers a robust and low-power option. However, for stakeholders such as municipal infrastructure managers and IoT device vendors, the faster convergence and reduced deployment time of Transformer-based models can enable quicker system updates and lower operational costs. %These results imply that for edge deployment scenarios prioritizing energy efficiency over marginal accuracy gains, LSTM offers a robust and cost-effective choice. However, when considering development time and automation scalability, Transformer-based models remain attractive due to their faster training convergence and end-to-end execution.

To further contextualize our approach, we compare against both Transformer and LSTM models from~\cite{singh2024data}, which perform forecasting on an NVIDIA A100 GPU (40GB VRAM). Their Transformer model reports an MSE of 0.62, 1.08~ms inference time, and a 184.59~MB model size, while their LSTM achieves an MSE of 0.63, 0.80~ms latency, and a 160~KB model size. In contrast, our work targets single-step forecasting under severe resource constraints, deploying quantized Transformer (19.84~KB) and LSTM (1.17~KB) models entirely on a battery-powered AMD Spartan-7 FPGA costing only 30 dollars. While simpler, this formulation shows that hardware–software co-design enables reliable local inference with minimal energy, essential for low-maintenance sensing infrastructure. %While our task formulation is simpler, the ability to achieve reliable local inference with minimal energy and memory highlights the value of hardware-software co-design for edge intelligence.

%%%%%%%%%%%%%%%%%%%%%%%%%%%%%%%%%%%%%%%%%%%%%%%%%%%%%%%%%%%%%%%%%%%%%%%%%%%%%%%%
\section{Conclusion and Future Work}
\label{sec:conclusion_future_work}

Our study presents an end-to-end forecasting framework for combined sewer overflow management, targeting overflow basin monitoring and forecastings. By deploying lightweight Transformer and LSTM models, compressed via integer-only quantization, on embedded FPGAs, our system enables fully local inference without reliance on cloud connectivity. The proposed framework satisfies key deployment requirements, including long-term battery-powered operation and tight hardware constraints, while offering flexible trade-offs between accuracy and energy consumption via automated hardware-aware model configuration optimization. Extensive evaluation on multi-year municipal datasets confirms the effectiveness of our approach in real-world conditions.
In future work, we will integrate our FPGA-accelerated edge deployment with the cloud-based solution in \cite{singh2025robust} into a unified hybrid system, coordinating local inference with cloud modeling for cross-site monitoring, adaptive updates, and smart city integration.

%\textcolor{red}{In future work, we will extend the evaluation beyond a single FPGA platform and dataset to assess portability across different architectures, sensing modalities, and deployment scales.}

%In future, we plan to improve training efficiency of quantized LSTM models and further optimize Transformer parallelism to enhance FPGA performance, enabling broader deployment across smart city infrastructure.

%%%%%%%%%%%%%%%%%%%%%%%%%%%%%%%%%%%%%%%%%%%%%%%%%%%%%%%%%%%%%%%%%%%%%%%%%%%%%%%%
\bibliographystyle{IEEEtran}
\bibliography{reference}

@article{saddiqi2023smart,
  title={Smart management of combined sewer overflows: From an ancient technology to {Artificial Intelligence}},
  author={Saddiqi, M Matin and Zhao, Wanqing and Cotterill, Sarah and Dereli, Recep Kaan},
  journal={Wiley Interdisciplinary Reviews: Water},
  year={2023},
}

@inproceedings{qian2023elasticai,
  title={{ElasticAI}: creating and deploying energy-efficient deep learning accelerator for pervasive computing},
  author={Qian, Chao and Ling, Tianheng and Schiele, Gregor},
  booktitle={International Conference on Pervasive Computing and Communications Workshops and other Affiliated Events},
  pages={297--299},
  year={2023},
  organization={IEEE}
}

@article{rosin2021committee,
  title={A committee evolutionary neural network for the prediction of combined sewer overflows},
  author={Rosin, TR and Romano, M and Keedwell, E and Kapelan, Z},
  journal={Water Resources Management},
  volume={35},
  number={4},
  year={2021},
  publisher={Springer}
}

@article{hammond2021detection,
  title={Detection of untreated sewage discharges to watercourses using {Machine Learning}},
  author={Hammond, Peter and Suttie, Michael and Lewis, Vaughan T and Smith, Ashley P and Singer, Andrew C},
  journal={NPJ Clean Water},
  volume={4},
  number={1},
  pages={18},
  year={2021},
  publisher={Nature Publishing Group UK London}
}

@inproceedings{wen2022transformers,
author = {Wen, Qingsong and Zhou, Tian and Zhang, Chaoli and Chen, Weiqi and Ma, Ziqing and Yan, Junchi and Sun, Liang},
title = {Transformers in time series: A survey},
year = {2023},
booktitle = {Proceedings of the Thirty-Second International Joint Conference on Artificial Intelligence},
}

@inproceedings{jacob2018quantization,
  title={Quantization and training of {Neural Networks} for efficient integer-arithmetic-only inference},
  author={Jacob, Benoit and Kligys, Skirmantas and Chen, Bo and Zhu, Menglong and Tang, Matthew and Howard, Andrew and Adam, Hartwig and Kalenichenko, Dmitry},
  booktitle={Proceedings of the IEEE conference on computer vision and pattern recognition},
  year={2018}
}

@inproceedings{ling2024towards,
  title={Towards auto-building of embedded {FPGA}-based soft sensors for wastewater flow estimation},
  author={Ling, Tianheng and Qian, Chao and Schiele, Gregor},
  booktitle={Annual Congress on Artificial Intelligence of Things},
  year={2024},
  organization={IEEE}
}

@article{nagel2021white,
  title={A white paper on {Neural Network} quantization},
  author={Nagel, Markus and Fournarakis, Marios and Amjad, Rana Ali and Bondarenko, Yelysei and Van Baalen, Mart and Blankevoort, Tijmen},
  journal={arXiv preprint arXiv:2106.08295},
  year={2021}
}

@article{shuvo2022efficient,
  title={Efficient acceleration of {Deep Learning} inference on resource-constrained edge devices: A review},
  author={Shuvo, Md Maruf Hossain and Islam, Syed Kamrul and Cheng, Jianlin and Morshed, Bashir I},
  journal={Proceedings of the IEEE},
  volume={111},
  number={1},
  year={2022},
  publisher={IEEE}
}

@article{qian2025configuration,
  title={Configuration-aware approaches for enhancing energy efficiency in {FPGA}-based {Deep Learning} accelerators},
  author={Qian, Chao and Ling, Tianheng and Cichiwskyj, Christopher and Schiele, Gregor},
  journal={Journal of Systems Architecture},
  year={2025},
  publisher={Elsevier}
}

@article{zhang2023review,
  title={A review of {Artificial Intelligence} in embedded systems},
  author={Zhang, Zhaoyun and Li, Jingpeng},
  journal={Micromachines},
  volume={14},
  number={5},
  pages={897},
  year={2023},
  publisher={MDPI}
}

@article{alshami2024iot,
  title={{IoT} Innovations in Sustainable Water and Wastewater Management and Water Quality Monitoring: A Comprehensive Review of Advancements, Implications, and Future Directions.},
  author={Alshami, Ahmad and Ali, Eslam and Elsayed, Moustafa and Eltoukhy, Abdelrahman EE and Zayed, Tarek},
  journal={IEEE Access},
  year={2024},
  publisher={IEEE}
}

@inproceedings{singh2024data,
  title={Data-driven Modeling of Combined Sewer Systems for Urban Sustainability: An Empirical Evaluation},
  author={Singh, Vipin and Ling, Tianheng and Chiaburu, Teodor and Biessmann, Felix},
  booktitle={The 47th German Conference on AI (2nd Workshop on Public Interest AI)},
  year={2024},
  organization={CEUR Workshop Proceedings},
  volume={3958}
}

@inproceedings{singh2025robust,
  title={Evaluating Time Series Models for Urban Wastewater Management: Predictive Performance, Model Complexity and Resilience (in proceedings)},
  author={Singh, Vipin and Ling, Tianheng and Chiaburu, Teodor and Biessmann, Felix},
  booktitle={The 10th International Conference on Smart and Sustainable Technologies},
  year={2025},
  organization={IEEE}
}

@article{allard2021climate,
  title={Climate change adaptation: Infrastructure and extreme weather},
  author={Allard, Ryan F},
  journal={Industry, Innovation and Infrastructure},
  pages={105--116},
  year={2021},
  publisher={Springer}
}

@article{van2022towards,
  title={Towards the long term implementation of real time control of combined sewer systems: A review of performance and influencing factors},
  author={Van Der Werf, Job Augustijn and Kapelan, Zoran and Langeveld, Jeroen},
  journal={Water Science and Technology},
  year={2022},
  publisher={IWA Publishing}
}

@article{sojobi2022impact,
  title={Impact of sewer overflow on public health: A comprehensive scientometric analysis and systematic review},
  author={Sojobi, Adebayo Olatunbosun and Zayed, Tarek},
  journal={Environmental research},
  volume={203},
  pages={111609},
  year={2022},
  publisher={Elsevier}
}

@inproceedings{baneerjee2024overflow,
  title={Overflow Prevention and Wastewater Harmony: Innovative Strategies for Urban Drain Management},
  author={Baneerjee, Avirup and Ranjan, Harsh and Debdas, Subhra and Srivastava, Ansh and Pandey, Aditya and Goyal, Sujal},
  booktitle={2024 IEEE 3rd World Conference on Applied Intelligence and Computing (AIC)},
  year={2024},
  organization={IEEE}
}

@article{qian2024exploring,
  title={Exploring energy efficiency of {LSTM} accelerators: A parameterized architecture design for embedded {FPGAs}},
  author={Qian, Chao and Ling, Tianheng and Schiele, Gregor},
  journal={Journal of Systems Architecture},
  volume={152},
  pages={103181},
  year={2024},
  publisher={Elsevier}
}

@inproceedings{ling2024integer,
  title={Integer-only Quantized {Transformers} for Embedded {FPGA}-based Time-series Forecasting in {AIoT}},
  author={Ling, Tianheng and Qian, Chao and Schiele, Gregor},
  booktitle={Annual Congress on Artificial Intelligence of Things},
  pages={38--44},
  year={2024},
  organization={IEEE}
}

@inproceedings{ling2025deployment,
  title     = {Automating Versatile Time-Series Analysis with Tiny {Transformers} on Embedded {FPGAs}},
  author    = {Ling, Tianheng and Qian, Chao and Haßler, Lukas and Schiele, Gregor},
  booktitle = {Proceedings of the 2025 IEEE Computer Society Annual Symposium on VLSI (ISVLSI)},
  year      = {2025},
  note      = {To appear},
  organization = {IEEE}
}

@article{nagpal2024optimizing,
  author = {Nagpal, Mudita and Siddique, Miran Ahmad and Sharma, Khushi and Sharma, Nidhi and Mittal, Ankit},
  title = {Optimizing wastewater treatment through {Artificial Intelligence}: Recent advances and future prospects},
  journal = {Water Science and Technology},
  volume = {90},
  number = {3},
  pages = {731-757},
  year = {2024},
  month = {07},
  issn = {0273-1223}
}

@article{monday2024review,
  author = {Monday, Celestine and Zaghloul, Mohamed S. and Krishnamurthy, Diwakar and Achari, Gopal},
  title = {A Review of {AI-Driven} Control Strategies in the Activated Sludge Process with Emphasis on Aeration Control},
  journal = {Water},
  volume = {16},
  year = {2024},
  number = {2},
  article-number = {305}
}

@article{seshan2024lstmencoder,
    author = {Seshan, Siddharth and Vries, Dirk and Immink, Jasper and van der Helm, Alex and Poinapen, Johann},
    title = {{LSTM}-based {Autoencoder} models for real-time quality control of wastewater treatment sensor data},
    journal = {Journal of Hydroinformatics},
    year = {2024},
}

@inproceedings{jiang2022ptq,
  author={Jiang, Hangyang and Li, Quande and Li, Yanteng},
  booktitle={2022 14th International Conference on Computer Research and Development (ICCRD)}, 
  title={Post Training Quantization after {Neural Network}}, 
  year={2022},
  volume={},
  number={},
  pages={1-6},
}

@Inbook{negi2021fpga,
  author="Negi, Anvit and Raj, Sumit and Thapa, Surendrabikram and Indu, S.",
  title="Field Programmable Gate Array (FPGA) Based IoT for Smart City Applications",
  bookTitle="Data-Driven Mining, Learning and Analytics for Secured Smart Cities: Trends and Advances",
  year="2021",
  publisher="Springer International Publishing",
  address="Cham",
  pages="135--158",
}
\end{document}